\documentclass[]{article}
\usepackage{nips12submit_e,times}

\usepackage{amssymb, amsmath, multirow,graphicx,subfigure,algorithm,algorithmic,amssymb,float,caption}

\nipsfinalcopy
\begin{document}

\title{Behavior Pattern Recognition using A New Representation Model}

\author{
Qifeng Qiao\\
Department of Systems Engineering\\
University of Virginia\\
Charlottesville, VA 22903 \\
\texttt{qq2r@virginia.edu} \\
\And
Peter A. Beling \\
Department of Systems Engineering\\
University of Virginia\\
Charlottesville, VA 22903 \\
\texttt{pb3a@virginia.edu} 
}
\maketitle

\begin{abstract}
We study the use of inverse reinforcement learning (IRL) as a tool for the recognition of agents' behavior on the basis of observation of their sequential decision behavior interacting with the environment.  We model the problem faced by the agents as a Markov decision process (MDP) and model the observed behavior of the agents in terms of forward planning for the MDP.  We use IRL to learn reward functions and then use these reward functions as the basis for clustering or classification models.   Experimental studies with \textsl{GridWorld}, a navigation problem,  and the \textsl{secretary problem}, an optimal stopping problem, suggest reward vectors found from IRL can be a good basis for behavior pattern recognition problems.  Empirical comparisons of our method with several existing IRL algorithms and with direct methods that use feature statistics observed in state-action space suggest it may be superior for behavior recognition problems. 
\end{abstract}

\section{Introduction}
The availability of sensing technologies, such as digital cameras, global position system, infrared sensors and others, makes the computer easily access the data recording the interaction between the agents and the environment. The new web technology also provides a large amount of background knowledge that describes the user behavior on the internet. More recent research has begun to build the behavior recognition system using the real-world data to understand the users' behavior.

Many approaches have been proposed to understand the agents' goals/plans from the observation of the decision behavior. The entire trace of actions is to be recognized and matched against a plan library or a set of possible goals/plans.  Despite of the success of these methods, they assume that the plan library, a set of possible goals or some behavior model are known beforehand and provided as an input. Goal information is often completely unknown in practice, however, and so it is difficult to model goals accurately.

Consider some examples in the real-world. Human behavior contains more complex structure and relations. E.g. Kautz pointed out two basic structure for behavior: decomposition and abstraction \cite{kautz1987}. Behavior can be decomposed into several events. For behavior recognition, Israeli security systems evaluate a series of events to reach a conclusion.  Are they merely loitering in the area? Are they wearing a warm coat on a hot day?  When a list of behavior hits a certain number, the system recognizes a  potential threat.  This method highly depends on the empirical experience and hardly recognizes the precise goal of the observed agent. To recommend personalized advertisement, the web companies hope to understand users' interest by analyzing the web browsing history. Are users interested in cameras, if they have booked a hotel recently? There is considerable interest in categorizing the users according to their interest, while it is more difficult to infer the precise goals of the users.  Due to the time and space limitation for advertisement, effective identification of the user's interest is of high importance for the companies. Another motivation for a new problem comes from domains like high frequency trading of stocks and commodities, where there is considerable interest in identifying new market players and algorithms based on observations of trading actions, but little hope in learning the precise strategies employed by these agents \cite{yang2012, paddrik2012}.

In this paper, we propose a new problem, termed {\emph{Behavior Pattern Recognition}}(BPR), that involves recognizing agents based on observation of their behavior in a sequential decision making setting. Broadly, the problem is to classify or cluster the agents according to the patterns of decision behavior that are learned from the samples of agents' decision-making process. The recognition problem  can be framed as a classification or clustering problem: (1) Given observations of decision trajectories consisting of sequential actions and given a label for each trajectory indicating which behavior pattern that agent has, the problem is to determine the behavior pattern for an agent with unlabeled trajectories. (2) Given only observations of the decision trajectories, the problem is to assign trajectories to clusters on the basis of similarity of behavior patterns.

A direct solution to BPR problem is to program some heuristic rules to recognize the behavior by decomposing complex behavior into a series of simple events and then evaluating them to reach a conclusion. However, programming the rules is hard. In contrast to the manually coded rules, we propose a learning model for BPR problem, characterizing the decision behavior with high level features in terms of the underlying goals.  Consider the problem of image recognition as an illustration of behavior recognition. In that problem, a computer learns to categorize images by representing every image as a multi-dimensional feature vector that consists of the components such as RGB color, texture, shape parameters or other advanced metrics. The key point of characterizing the behavior is how to effectively find a high level vector that represents the sequential behavior and encodes the information on patterns. From the perspective of decision-making process, the underlying goal of an agent is considered as an abstract representation of the behavior. If a decision-making process is modeled by MDP, the reward function is assumed to encode the goal of that agent.  

IRL \cite{andrew2000} addresses the task of learning a reward function for given MDP that is consistent with observations of optimal decision making for the process. An assumption is that the expert's goal/intention can be characterized by the reward function. If the expert is rational, the demonstration behavior should aim to maximize the long-term accumulative reward. We study the use of IRL to characterize the decision behavior, modeling  the problem faced by the agents as a MDP and assuming the reward function of the MDP model as a high-level abstraction of the decision behavior. The motivation is that even when the true behavior is not rational and we can't learn the precise goals/decision-strategies,  we still can categorize the agents by learning the reward functions that make MDP models approximate the observed behavior.

IRL has received increasing attention in the machine learning field in recent years. Most of this work is focused on apprenticeship learning, in which IRL is used as the core method for finding decision policies consistent with observed behavior \cite{abeel2004, neu2007, umar2008b}. A number of  IRL algorithms and modeling constructs have been proposed  for apprenticeship learning or imitation learning, including Max-margin planning \cite{ratliff06}, gradient tuning methods \cite{neu2007}, linear solvable MDP \cite{dvijotham2010}, bootstrap learning \cite{boularias2010}, feature construction \cite{levine2010}, Gaussian process IRL \cite{qiao2011} and Bayesian inference \cite{choi2011}. 

On two well-know sequential decision-making problems, we compare our method with several existing IRL algorithms and with direct methods that use feature statistics observed in state-action space.  Our main contributions include: (1) identification of a new learning task that categorizes agents by learning their behavior patterns; (2) design of simple methods to solve the BPR problem that characterize behavior in original observation space; (3) development of a new model-based method to solve the BPR problem in MDP reward space; and (4) observation that our new method using reward space provides a formal way to solve the behavior recognition problem and performs superior to other methods.
 
\section{Preliminaries}
We define the input of BPR problem as a tuple $\mathbf{B} = (D_{1}, D_{2}, \ldots\, D_{N})$, where $D_{n}, n\in\{1,2,\ldots,N\}$ is the observation of the $n-th$ agent. For a classification problem, $D_{n}=(\mathcal{O}_{n},y_{n})$, where $\mathcal{O}_{n}$ is a set of observed decision trajectories and $y_{n}$ is the class label for the $n-th$ agent. The agents, who have the same behavior patterns, are given the same class label. Similarly, in a clustering problem, $D_{n}$ only consists of the observed decision trajectories.

We define the set of decision trajectories $\mathcal{O}_{n}=\{h_{n}^{j}\},\ j=1,2,\ldots,|\mathcal{O}_{n}|$, where each trajectory $h_{n}^{j}$ is defined as a series of state and action pairs: $\{(s,a)_{n}^{t}\}, t=1,2,\ldots,|h_{n}^{j}|$. Here, the $s$ denotes the state for the decision problem and the $a$ means the action selected by the agent at state $s$.

To determine a label for an agent, we may develop a model to decompose the observed behavior into several events. Each event can be described by a complete or part of a decision trajectory. When a list of events hits a certain number, the model recognizes a label for an agent. However, this method requires a lot of domain knowledge and human experience to program the heuristic rules. 

Another way to solve BPR problem is to effectively represent the problem in a multi-dimensional space and then apply the learning algorithms. The decision-making process can be characterized in two layers. The outer layer characterizes the behavior by calculating some statistic information on the observed state and action. The inner layer is an abstraction of the behavior, which is related to the goal or the internal mind of the agent that determines the behavior fundamentally.

\section{Simple Representation of Behavior Recognition Problem}
In this section, we describe two methods in outer layer that categorize the decision-making agents just based on observation.

The first method is called feature trajectory (\textit{FT}). Assume the length of a decision trajectory is $H$. The vector to characterize the behavior in $j-th$ decision trajectory is written as follows.
\begin{eqnarray}
f(h_{n}^{j})=[s_{1}, a_{1}, s_{2}, a_{2},\ldots,s_{H},a_{H}],\nonumber
\end{eqnarray}
where $s_{i},i\in\{1,2,\ldots,H\}$ is a discrete random variable meaning the state index at $i-th$ decision stage, and $a_{i}$ represents the action selected at state $s_{i}$. E.g., we have a problem that can be defined by 3 states and 2 actions. Then $s_{i}\in\{1,2,3\}$ and $a_{i}\in\{1,2\}$. In the observation, every trajectory starts from the same initial state. Given the observation set $\mathcal{O}_{n}$ for $n-th$ agent, the feature vector $f_{n}$ is obtained by computing this equation: $f_{n}=\frac{1}{|\mathcal{O}_{n}|}\sum_{j=1}^{|\mathcal{O}_{n}|}f(h_{n}^{j})$, where the vector $f(h_{n}^{j})$ is preprocessed by scale-normalization before averaging.

Then, the $n-th$ agent is represented by a feature vector $f_{n}$. Consider a supervised learning problem. Given a real valued input vector $f_{n}\in\mathcal{F}$ and a category label $y_{n}\in\mathcal{Y}$, we aim to learn a function $h:\mathcal{F}\rightarrow\mathcal{Y}$. 

The second method is called feature expectation (\textit{FE}), which has been widely used by apprenticeship learning as a representation of the averaged long-term performance. Assume a basis function $\phi:\mathcal{S} \rightarrow [0,1]^{d}$, where $\mathcal{S}$ denotes the state space. The feature expectation $f_{n}=\frac{1}{|\mathcal{O}_{n}|}\sum^{|\mathcal{O}_{n}|}_{j=1}\sum_{s_{t}\in h_{n}^{j}}\gamma^{t}\phi(s_{t})$, where $\gamma\in(0,1)$ is a discount factor. The associated apprenticeship learning algorithms aim to find a policy that performs as well as demonstrations by minimizing the distance between their feature expectations.  Here, we only use the observed state sequence to compute the feature expectation vector for an agent, where the $\gamma$ is manually defined constant, e.g. $0.95$. Then, the $n-th$ agent can be represented by the vector $f_{n}$ that is obtained from $\mathcal{O}_{n}$.

\section{A New Representation Model}
To solve the BPR problem with high-level feature representation, we propose to use the following steps.
\begin{enumerate}
\item Given the BPR problem with input $\mathbf{B}$, we use the set $\{\mathcal{O}_{n}\}, n\in\{1,2,\ldots,N\}$ to construct the state space $\mathcal{S}$ and action space $\mathcal{A}$ for the decision-making problem.
\item  
For $n-th$ observed agent, we assume an MDP model $M=(\mathcal{S}, \mathcal{A}, R_{n}, \gamma, \mathcal{P})$, where $R_{n}$ is the unknown reward function for this agent, $\gamma$ is the constant discount factor, and $\mathcal{P} = \left\{\mathbf{P}_{a}\right\}_{a\in \mathcal{A}}$ is a set of  transition probability matrices $\mathbf{P}_{a}$ for action $a\in \mathcal{A}$. The entries of $\mathbf{P}_{a}$, written as $\mathbf{P}_{a}(s,s')$, give the probability of transitioning to state $s'\in \mathcal{S}$ from state $s\in\mathcal{S}$ given the action is $a$. The rows of $\mathbf{P}_{a}$, denoted $\mathbf{P}_{a}(s,:)$, give a probability vector of transitioning from state $s$ to all the states in $\mathcal{S}$. The $\mathcal{P}$ can be modeled using prior knowledge of the problem or estimated from the observed decision trajectories $\{\mathcal{O}_{n}\}$. In a finite state space the reward function $R_{n}$ may be considered as a vector, $r_{n}$, whose elements give the reward in each state. Here we expect that there exists an unknown reward function that can make the MDP find a policy as similar as the observed behavior.
\item Apply IRL algorithms to learn the reward vector $r_{n}$ for $n-th$ agent. 
\item Estimate the reward vectors for every agent. Then the supervised learning problem is written as: given the real valued input vector $r_{n}\in \mathcal{R}$ and the category label $y_{n}\in \mathcal{Y}$, we aim to learn a function $h:\mathcal{R}\rightarrow \mathcal{Y}$. 
\item Given a new observed agent, we repeat step 1-3 to get the reward vector for the agent and then predict the label for the behavior pattern using estimated function $h:\mathcal{R}\rightarrow \mathcal{Y}$. 
\end{enumerate}

In MDP model, a policy is defined as a mapping $\pi:\mathcal{S}\rightarrow{A}$. The value function for a policy $\pi$ is $V^{\pi}(s_{0})=E[\sum^{\infty}_{t=0}\gamma^{t}R(s_{t})|p(s_{0}),\pi]$ where $p(s_{0})$ is the distribution of the initial state and the action at state $s_{t}$ is determined by policy $\pi$. Similarly, the Q function is defined as $Q(s,a) = R(s) + \gamma \sum_{s'\in\mathcal{S}} \mathbf{P}_{a}(s,s')V^{\pi}(s')$. At state $s$, an optimal action is selected by $a^{*}=\max_{a\in\mathcal{A}}Q(s,a)$.

An instance of the IRL problem is written as a triplet  $B = (M\setminus r, p(r), \mathcal{O})$, where $M\setminus r$ is a MDP model without the reward function and $p(r)$ is prior knowledge on the reward. The vector $p(r)$ can be a non-informative prior if we have no knowledge about the reward function or a Gaussian or other distribution  if we model the reward as a specific stochastic process.

We use an MDP to model the decision problem faced by an agent under observation.  The reality of the agent's decision problem and process may differ from the MDP model,  but we interpret every observed decision of the agent as the choice of an action in the MDP.  The dynamics of the environment in the MDP are described by the transition probabilities $\mathcal{P}$.  These probabilities may be interpreted as being a prior, if known in advance, or as an estimation of the agent's beliefs of the dynamics. Next, we will show how to learn the reward functions by employing some exiting IRL algorithms.

\section{Bayesian framework for IRL}
Most existing IRL algorithms assume that the agents are perfectly rational and the observed behavior is optimal.
 Prominent examples include  the model in \cite{andrew2000}, which we term linear IRL (\textit{LIRL}) because of its linear nature,   \textit{WMAL} in \cite{umar2008b}, and \textit{PROJ} in \cite{abeel2004}. In these algorithms, the reward function is written linearly in terms of features as 
$R(s) = \sum_{i=1}^{d}\omega_{i}\phi_{i}(s) = \omega^{T}\phi(s)$, where $\phi:\mathcal{S} \rightarrow [0,1]^{d}$ and $\omega^{T} = [\omega_{1},\omega_{2},\cdots, \omega_{d}]$.

Our computational framework uses Bayesian IRL to estimate the reward vectors in a MDP, which was initially proposed in \cite{deepak2007}. The posterior over reward function for $n-th$ agent is written as
\begin{eqnarray}
p(r_{n}|\mathcal{O}_{n})=p(\mathcal{O}_{n}|r_{n})p(r_{n}) \propto \prod^{|\mathcal{O}_{n}|}_{j=1} \prod_{(s,a)\in h_{n}^{j}} p(a|s,r_{n}).  \nonumber
\end{eqnarray}
Then, the IRL problem is written as $ \max_{r_{n}} \log p(\mathcal{O}_{n}|r_{n})+\log p(r_{n})$. For many problems, however, the computation of $p(r_{n}|\mathcal{O}_{n})$ may be complicated and some algorithms use Markov chain Monte Carlo (MCMC) to sample the posterior probability. Considering the computation complexity to deal with a large number of IRL problems, we choose the IRL algorithms that have well defined likelihood function to reduce the computation cost. 
 
 \subsection{IRL with Boltzmann Distribution}
 To model the likelihood function, some IRL algorithm in \cite{monica2011}, which we call maximum likelihood IRL (\textit{MLIRL}), uses Boltzmann distribution to calculate $p(a|s,r_{n})$ using $p(a|s,r_{n}) = \frac{e^{Q(s,a)}}{\sum_{a\in\mathcal{A}}e^{Q(s,a)}}$.
 
\subsection{IRL with Gaussian Process}
IRL algorithm, which is called \textit{GPIRL} in \cite{qiao2011}, uses preference relations to model the likelihood function $P(\mathcal{O}_{n}|r_{n})$ and assumes the $r_{n}$ is generated by Gaussian process for $n-th$ observed agent.

 Given a state, we assume that the optimal action is selected according to
Bellman optimality. The preference relation is defined as follows.
 
At state $s$, $\forall \hat{a}, \check{a} \in \mathcal{A}$, we define the \textsl{action preference relation} as: 
\begin{enumerate}
	\item Action $\hat{a}$ is weakly preferred to $\check{a}$, denoted as $\hat{a}\succeq_{s} \check{a}$, if $Q(s,\hat{a})\geq Q(s,\check{a})$;
	\item Action $\hat{a}$ is strictly preferred to $\check{a}$, denoted as $\hat{a}\succ_{s} \check{a}$, if $Q(s,\hat{a})> Q(s,\check{a})$;
	\item Action $\hat{a}$ is equivalent to $\check{a}$, denoted as $\hat{a}\sim_{s} \check{a}$, if and only if $\hat{a}\succeq_{s} \check{a}$ and $\check{a}\succeq_{s} \hat{a}$.
\end{enumerate}
 
Given the observation set $\mathcal{O}_{n}$, we have a group of preference relations at each state $s$, which is written as
\begin{eqnarray}
\mathcal{E} \equiv \left\{(\hat{a} \succ_{s} \check{a}),\ \hat{a}\in \hat{\mathcal{A}},\ \check{a}\in \mathcal{A}\setminus\hat{\mathcal{A}} \right\}
\cup \left\{(\hat{a}\sim_{s} \hat{a}'),\ \hat{a},\hat{a}'\in \hat{\mathcal{A}}\right\},\nonumber
\label{eq:prefset}
\end{eqnarray}
where $\hat{\mathcal{A}}\in\mathcal{A}$ is the action subspace for state $s$ obtained from the set $\mathcal{O}_{n}$.

Let $\textbf{r}$ be the vector of $r_{n}$ containing the reward for $m$ possible actions at $T$ observed states. We have
\begin{eqnarray}
\textbf{r}&=&(\underbrace{\textbf{r}_{a_{1}}(s_{1}),...,\textbf{r}_{a_{1}}(s_{T})}, \ldots ,\underbrace{\textbf{r}_{a_{m}}(s_{1}), \ldots ,\textbf{r}_{a_{m}}(s_{T})})\nonumber\\
&=& (\ \ \ \ \ \ \ \ \ \ \ \textbf{r}_{a_{1}},\ \ \ \ \ \ \ \cdots,\ \ \ \ \ \ \ \ \ \ \ \ \ \textbf{r}_{a_{m}}),\nonumber
\label{defonf}
\end{eqnarray}
 where $T=|\mathcal{S}|$ and $\textbf{r}_{a_{m}}, \forall m\in\{1,2,\ldots,|\mathcal{A}|\}$, denotes the reward with respect to $m$-th action.
  
Consider $\textbf{r}_{a_{m}}$ as a Gaussian process if, for any $\left\{s_{1},\cdots,s_{T}\right\} \in \mathcal {S}$, the random variables $\left\{\textbf{r}_{a_{m}}(s_{1}), \cdots, \textbf{r}_{a_{m}}(s_{T})\right\}$ are normally distributed. We denote by $k_{a_{m}}(s_{c}, s_{d})$ the function generating the value of entry $(c,d)$ for covariance matrix $\textbf{K}_{a_{m}}$, which leads to $\textbf{r}_{a_{m}}\sim N(0,\textbf{K}_{a_{m}})$. 
Then the joint prior probability of the reward is a product of multivariate Gaussian, namely 
$p(\textbf{r}|\mathcal{S})=\prod\nolimits^{|\mathcal{A}|}_{m=1}p(\textbf{r}_{a_{m}}|\mathcal{S})\,\ {\rm and}\ \textbf{r}\sim N(0, \textbf{K})$.  Note that $\textbf{r}$ is completely specified by the positive definite covariance matrix $\textbf{K}$. 

A simple strategy is to assume that the $|\mathcal{A}|$ latent processes are uncorrelated. Then the covariance matrix $\textbf{K}$ is block diagonal in the covariance matrices $\left\{\textbf{K}_{1},...,\textbf{K}_{|\mathcal{A}|}\right\}$. In practice, we use a squared exponential kernel function, written as:
\begin{eqnarray}
k_{a_{m}}(s_{c},s_{d}) = e^{\frac{1}{2}(s_{c}-s_{d})\textbf{M}_{a_{m}}(s_{c}-s_{d})}+\sigma^{2}_{a_{m}}\delta(s_{c},s_{d}),\nonumber
\end{eqnarray}
where  $\textbf{M}_{a_{m}}=\kappa_{a_{m}} \textbf{I}_{T}$ and $\textbf{I}_{T}$ is an identity matrix of size $T$. The function $\delta(s_{c},s_{d})=1$, when $s_{c}=s_{d}$; otherwise $\delta(s_{c},s_{d})=0$. Under this definition the covariance is almost unity between variables whose inputs are very close in the Euclidean space, and decreases as their distance increases. 

Then, the \textit{GPIRL} algorithm estimates the reward function by iteratively conducting the following two main steps:
\begin{enumerate}
\item Get estimation of $r_{MAP}$ by maximizing the posterior $p(r_{n}|\mathcal{O}_{n})$, which is equal to minimize $-\log p(\mathcal{O}_{n}|r_{n})-\log p(r_{n}|\theta)$, where $\theta = (\kappa_{a_{m}}, \sigma_{a_{m}})$ is the hyper-parameter controlling the Gaussian process, and $p(\mathcal{O}_{n}|r_{n})=\prod p((\hat{a}\succ_{s}\check{a})) \prod p((\hat{a} \sim_{s}\hat{a}'))$. Above optimization problem has been proved to be convex programming in \cite{qiao2011}.
\item Find the optimized hyper-parameters by applying gradient decent optimization method to maximize $\log p(\mathcal{O}_{n}|\theta, r_{MAP})$, which is the Laplace approximation of $p(\theta|\mathcal{O}_{n})$.
\end{enumerate}

\section{Experimentation}
\label{section:exp}
Our experiments are designed to evaluate the performance of IRL algorithms in behavior recognition in comparison to methods that use simple feature representations obtained directly from observation space.  We study two problems, {\em GridWorld} and the {\em secretary problem}. {\em GridWorld} provides insight into the task of recognizing machine agents for decision problems that may be modeled using MDP models under the strict rationality assumption.  The secretary problem provides an environment in which the agents do not act with respect to the solution of an MDP.   Agents in the secretary problem employ heuristic decision rules derived from experimental study of human behavior in psychology and economics. 

To evaluate the recognition performance, we use the following algorithms: (1) Clustering: Kmeans \cite{duda2001}; (2) Classification: Support vector machine (SVM), K-nearest neighbors (KNN), Fisher discriminant analysis (FDA) and logistic regression (LR) \cite{duda2001}. We use clustering accuracy \cite{xu2005} and Normalized Mutual Information (NMI) \cite{strehl2002} to compare clustering results.   

\subsection{{\em GridWorld} problem}
In the \textsl{GridWorld} problem, which is used as a benchmark experiment by Ng and Russell in \cite{andrew2000}, an agent  starts from a given square and moves towards a destination square. The agent has five actions to take: moving in the four cardinal directions or staying put. With probability 0.65 the agent moves to its chosen location, with probability 0.15  it stays in the same location regardless of chosen action, and with probability 0.2 it moves in a random cardinal direction. 

The IRL problem for \textsl{GridWorld} is to recover the reward structure given the observations of agent actions.  To produce these observations, we first simulate the agent's behavior using the optimal solution of an MDP to decide how to move in the \textsl{GridWorld}.  We then collect observation data by sampling the simulated movement. Note that the reward function of the MDP used for simulating the agents is not known to the IRL learner.

We investigated the  behavior recognition problem in terms of clustering and classification on  a $10\times 10$ \textsl{GridWorld} problem. Experiments were conducted according to the steps in Algorithm \ref{alg:exp}.

\begin{algorithm}
\footnotesize
\caption{{\em GridWorld} experimentation steps}
\label{alg:exp}
\begin{algorithmic}[1]
\STATE Input the variables $\mathcal{S}, \mathcal{A} \mbox{ and } \mathcal{P}$.  Design two ground truth reward functions written as $r^{*}_{1}$ and $r^{*}_{2}$. 
\STATE Simulate  agents and sample their behavior.
  \FOR{$i=1 \to 2$}
  \FOR{$j=1 \to 200$}
  \STATE Model an agent using $M=(\mathcal{S}, \mathcal{A}, \mathcal{P}, r_{ij},\gamma)$, where the reward $r_{ij} = r^{*}_{i}+ \mbox{random Gaussian noise}$.
  \STATE Sample  decision trajectories $\mathcal{O}_{ij}$, and make the ground truth label $y_{ij}=0$, if $i=1$; $y_{ij}=1, \mbox{if } i=2$.
  \ENDFOR
  \ENDFOR 
\STATE IRL has access to the problem $B=(\mathcal{S}, \mathcal{A}, \mathcal{P}, \gamma, \mathcal{O}_{ij})$ for this agent, and then infers the reward $r_{ij}$.
\STATE Recognize these agents based on the learned $r_{ij}$.
\end{algorithmic}
\end{algorithm}
The simulated agents in our experiments have hybrid destinations. A small number of short decision trajectories tends to present challenges to action feature methods, which is an observation of particular interest. Additionally, the length of trajectories may have a substantial impact on performance. If the length is so long that the observed agent reaches the destination in every trajectory, the problem can be easily solved based on observations. Thus, we evaluate and compare performance by making the length of decision trajectory small.

\begin{minipage}{\textwidth}
\begin{minipage}[b]{0.49\textwidth}
\small
\centering
\begin{tabular*}{\textwidth}{@{\extracolsep{\fill}}c|rrrrr|}
\hline
$|\mathcal{O}_{n}|$& FE &       FT &   PROJ &      GPIRL  \\
\hline
        4 &     0.0077 &     0.0012 &     0.0068 &     0.0078  \\

        8 &     0.0114 &     0.0016 &     0.0130 &     0.0932  \\

        16 &     0.0177 &     0.0014 &     0.0165 &     0.7751  \\

        20 &     0.0340 &     0.0573 &     0.0243 & { 0.8113}  \\

        30 &     0.0321 &     0.0273 &     0.0365 & { 0.8119}  \\

       40 &     0.0361 &     0.0459 &     0.0389 & { 0.8123} \\

       60 &     0.0387 &     0.0467 &     0.0388 & { 0.8149}  \\

       80 &     0.0441 &     0.1079 &     0.0421 & { 0.8095}  \\

       100 &     0.0434 &     0.1277 &     0.0478 & { 0.8149} \\

       200 &     0.0502 &     0.1649 &     0.0498 & { 0.8149}  \\

\hline
\end{tabular*}  
\label{table:nmi}
\captionof{table}{NMI scores for \textsl{GridWorld} problem}
\end{minipage}
\hfill
\begin{minipage}[b]{0.49\textwidth}
	\centering
	  \includegraphics[width =2.4in]{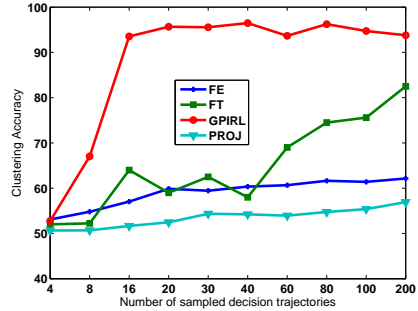}
	  \captionof{figure}{Clustering accuracy}
\label{fig:clusteracc}
\end{minipage}
\end{minipage}

\begin{figure}[!htb]
	\centering
	\subfigure[SVM]
	{
	  \includegraphics[width =2.6in]{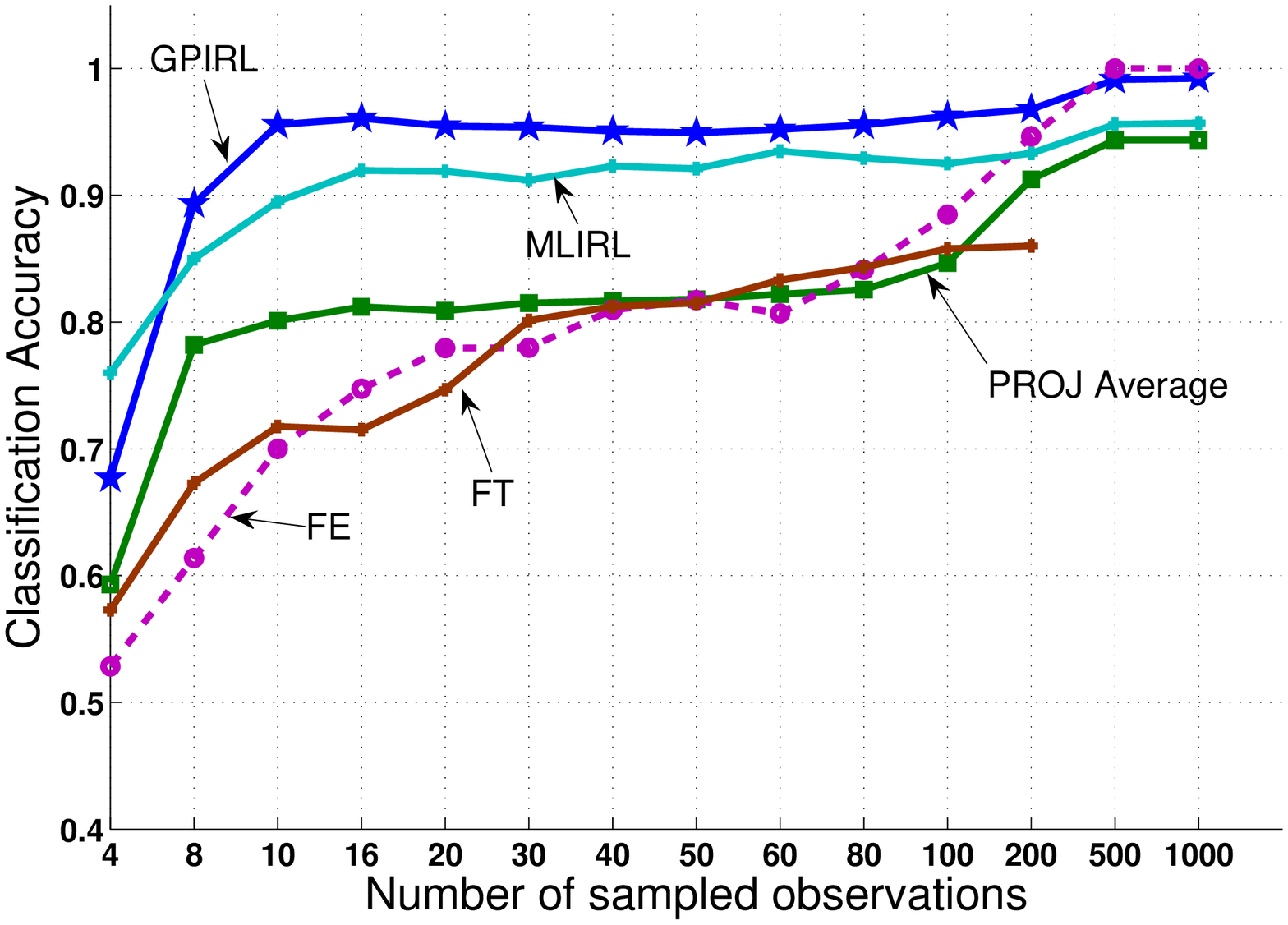}
	}
	\hfil
	\subfigure[KNN]
	{
	  \includegraphics[width =2.6in]{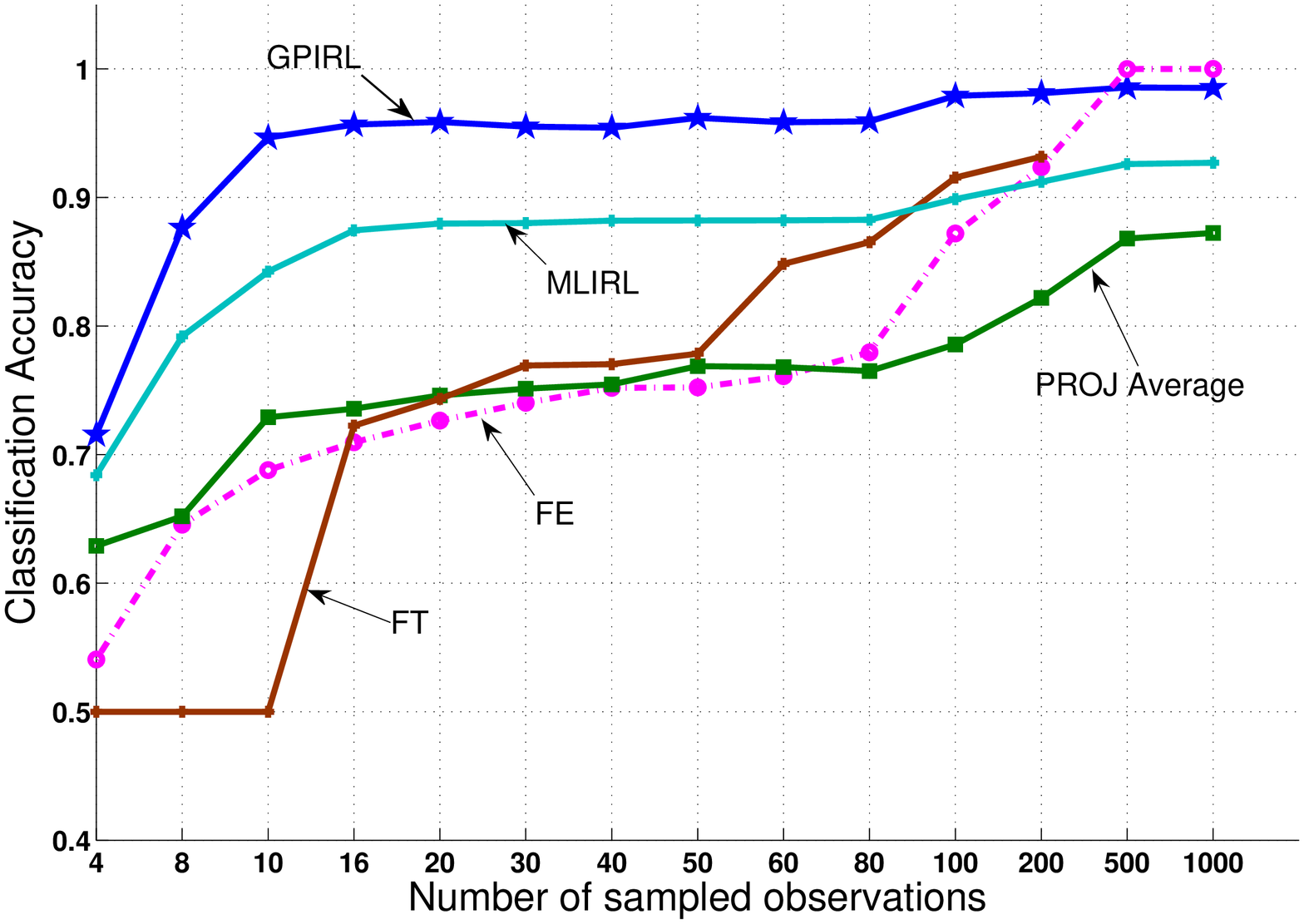}
	}
	\subfigure[LR]
	{
	  \includegraphics[width =2.6in]{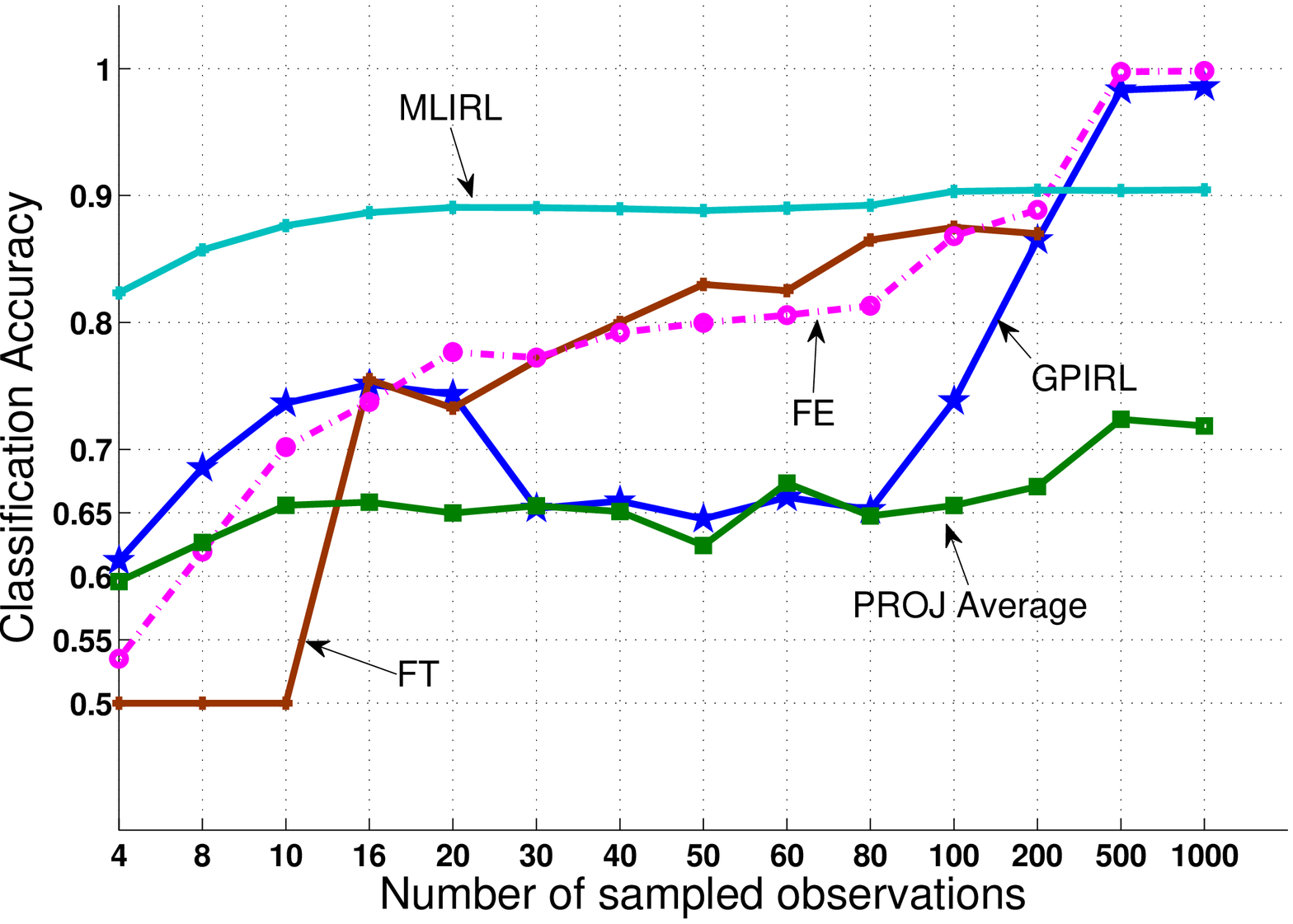}
	}
	\hfil
	\subfigure[FDA]
	{
	  \includegraphics[width =2.6in]{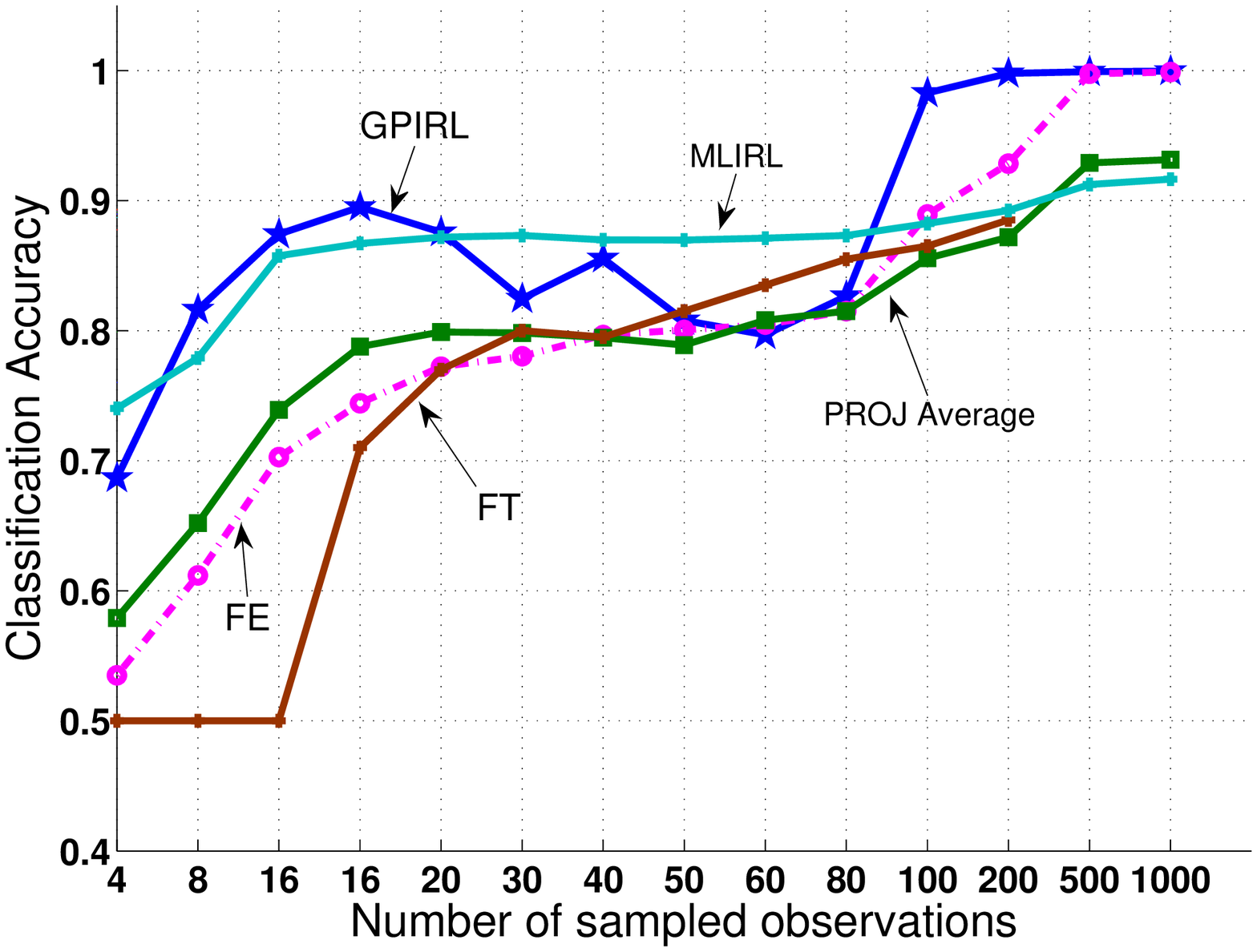}
	}
	\caption{Classification results with respect to different classifier.}
\label{fig:classifygridsvm}
\end{figure}
Table \ref{table:nmi} displays NMI scores and  \figurename \ref{fig:clusteracc} shows clustering accuracy. The length of the trajectory is limited to six steps, as we assume the observation is incomplete and the learner does not have sufficient information to infer the goal directly. Results are averaged over 100 replications. Clustering performance improves with increasing number of observations. When the number of observations is small, \textsl{GPIRL} method achieves high clustering accuracy and NMI scores due to the advantage of finding more accurate reward functions that can well characterize the decision behavior.  The IRL algorithms, such as \textsl{PROJ} and \textsl{WMAL}, are not effective in this problem because the length of the observed decision trajectory is too small to provide a feature expectation that is a good approximation to the agent's long-term goals. Considering the utilization of feature learning algorithms to improve the simple feature representations, we also did experiments with PCA-based features where the projection sub-space is spanned by those eigenvectors that correspond to the principal components $c = 10, 20,\ldots,90$ for \textit{FE} and $c=2,4,6,8,10$ for \textit{FT}. No significant changes in the clustering NMI scores and accuracy scores are observed. Therefore, we do not show the performance of PCA-based features in Table \ref{table:nmi}and Figure \ref{fig:clusteracc}.

 \figurename \ref{fig:classifygridsvm} displays  classification accuracy for a binary classification problem in which there are four hundred agents coming from two groups of decision strategies. The results are averaged over 100 replications with tenfold cross-validations. Four popular classifiers (SVM, KNN, FDA and LR) are employed to evaluate the classification performance. Results suggest that the classifiers based on IRL perform better than the simple methods, such as \textit{FT} and \textit{FE}, particularly when the number of observed trajectories and the length of the trajectory are small.  The results support our hypothesis that recovered reward functions constitute an effective and robust feature space for clustering or classification analysis in a behavior pattern recognition setting. 

\subsection{Secretary problem}
\begin{algorithm}
\small
\caption{Experimentation with Secretary Problem}
\label{alg:sec}
\begin{algorithmic}[1]
\STATE Given a heuristic rule with a parameter $h$, $k$ or $\ell$. \STATE Add random Gaussian noise to the parameter, which is written as $\hat{p}$. 
\STATE Generate new secretary problem with $X$ applications and let $n-th$ agent solve these problems using this heuristic rule with its own parameter $\hat{p}$. Save the observed decision trajectories into $\mathcal{O}_{n}$.
\STATE Model the secretary problem in terms of an MDP consisting of the following components:
\begin{enumerate}
\item State space $\mathcal{S}=\{1,2,\ldots, X\}$, where $s\in\mathcal{S}$ means that at time $s$ the current applicant is a candidate. 
\item Action space $\mathcal{A}$ consisting of two actions: reject and accept.
\item Transition probability $\mathcal{P}$, computed as follows:  given the reject action, the probability of transitioning from state $s_{i}$ to $s_{j}$, $p(s_{j}|s_{i})$, is $\frac{s_{i}}{s_{j}(s_{j}-1)}$ if $s_{j}\geq s_{i}$, and $0$ otherwise; given the accept action, the probability of transitioning from state $s_{i}$ to $s_{j}$, $p(s_{j}|s_{i})$, is 1 if $s_{i}=s_{j}$, and $0$ otherwise.
\item The discount factor $\gamma$ is a selected constant.
\item The reward function is unknown.
\end{enumerate}
\STATE Infer the reward function by solving an IRL problem  $B=(\mathcal{S}, \mathcal{A}, \mathcal{P}, \gamma, \mathcal{O}_{n})$.
\end{algorithmic}
\end{algorithm}
The secretary problem is a sequential decision-making problem in which the binary decision to either stop or continue a search is made on the basis of objects already seen.  As suggested by the name, the problem is usually cast in the context of interviewing applicants for a secretarial position. The decision maker interviews a randomly-ordered sequence of applicants one at a time.  The applicant pool is such that the interviewer can unambiguously rank each applicant  in terms of quality relative to the others seen up to that point.  After each interview, the decision maker chooses either to move on to the next applicant, forgoing any opportunity to hire the current applicant, or to hire the current applicant, which terminates the process.  If the process goes as far as the final applicant, he or she must be hired.  Thus the decision maker chooses one and only one applicant.  The objective is to maximize the probability that the accepted applicant is, in fact, the best in the pool. 

To test our hypotheses on BPR, an idea experiment would involve recognizing individual human decision makers on the basis of observations of hiring decisions that they make in secretary problem simulations.  Experiments with human decision making for the secretary problem are reported on in \cite{seale1997, schunk2009},   but raw data consisting of decision maker action trajectories is not available.  However, a major conclusion of these studies is that the decisions made by the humans largely can be explained in terms of three decision strategies, each of which uses the concept of a {\em candidate}.  An applicant is said to a candidate he or she is the best applicant seen so far.  The decision strategies of interest are the:

\begin{enumerate}
\item {\em Cutoff rule (CR)} with cutoff value $h$, in which the agent will reject the first $h-1$ applicants and  accept the next candidate; 
\item {\em Successive non-candidate counting rule (SNCCR)} with parameter value $k$, in which the agent will accept the first candidate who follows $k$ successive non-candidate applicants since the last candidate; and
\item  {\em Candidate counting rule (CCR)} with parameter value $\ell$, in which  the agent selects the next candidate once $\ell$ candidates have been seen. 
\end{enumerate}

The optimal decision strategy for the secretary problem is to use CR with a parameter that can be computed using dynamic programming for any value of $n$, the number of secretaries.  As $n$ grows, the optimal parameter converges to $n/e$ and yields a probability of successfully choosing the best applicant that converges to $1/e$.  Thus only one of the three decision strategies enumerated above can be viewed as optimal, and that only for a single parameter value out of the continuum of possible values.  Human actions are usually suboptimal and tend to look like mixtures of CR (with a non-optimal parameter), SNCCR, and CCR \cite{seale1997}.

As a surrogate for the action trajectories of humans, we use agents that we generate action trajectories for randomly sampled secretary problems using CR, SNCCR, and CCR.  For a given decision rule (CR, SNCCR, CCR), we simulate a group of agents that adopt this rule, differentiating individuals in a group by adding Gaussian noise to the rule's parameter.  The details of the process are given in Algorithm \ref{alg:sec}.  We use IRL and observed actions to learn reward functions for  the MDP model given in Algorithm \ref{alg:sec}. It is critical to understand that the state space for this MDP model captures nothing of the history of candidates, and as a consequence is wholly inadequate for the purposes of modeling SNCCR and CCR.  In other words, for general parameters, neither SNCCR nor CCR can be expressed as a policy for the MDP in Algorithm \ref{alg:sec}.  (There does exist an MDP in which all three of the decision rules can be expressed as policies, but the state space for this model is exponentially larger.) Hence, for two of the rules, the processes that we use to generate data and the processes we use to learn are distinct.

As an initial set of experiments, we generated an equal number of agents from each rule. All the heuristic rules use the same parameter value. We have compared the method using statistical feature representations obtained from the raw decision trajectories and our IRL model-based method. We employ 10 fold cross-validation to obtain the average accuracy, and it is always 100\% . 

\begin{figure} 
\centering
\includegraphics[width=2.5in]{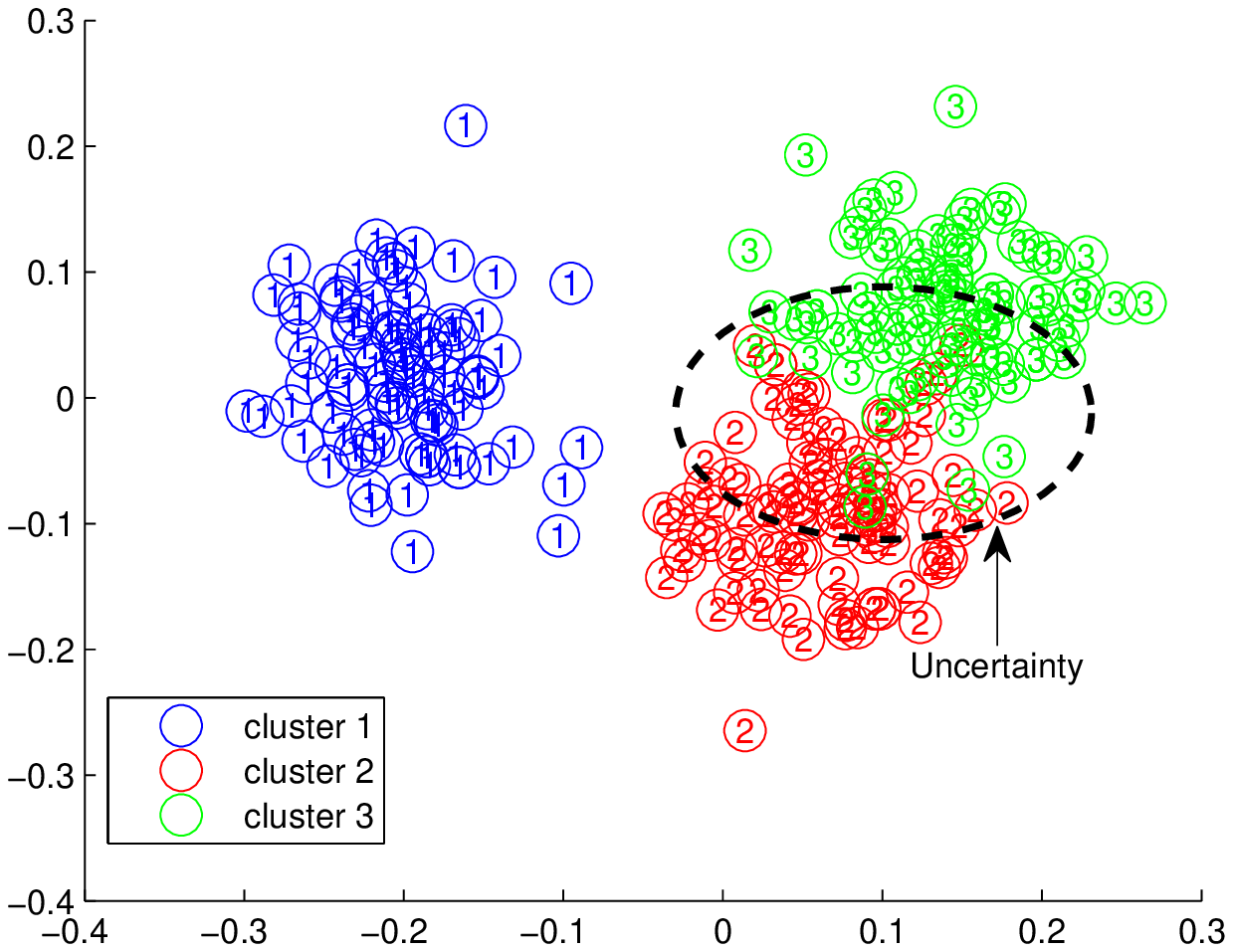}
\hfill
\includegraphics[width=2.5in]{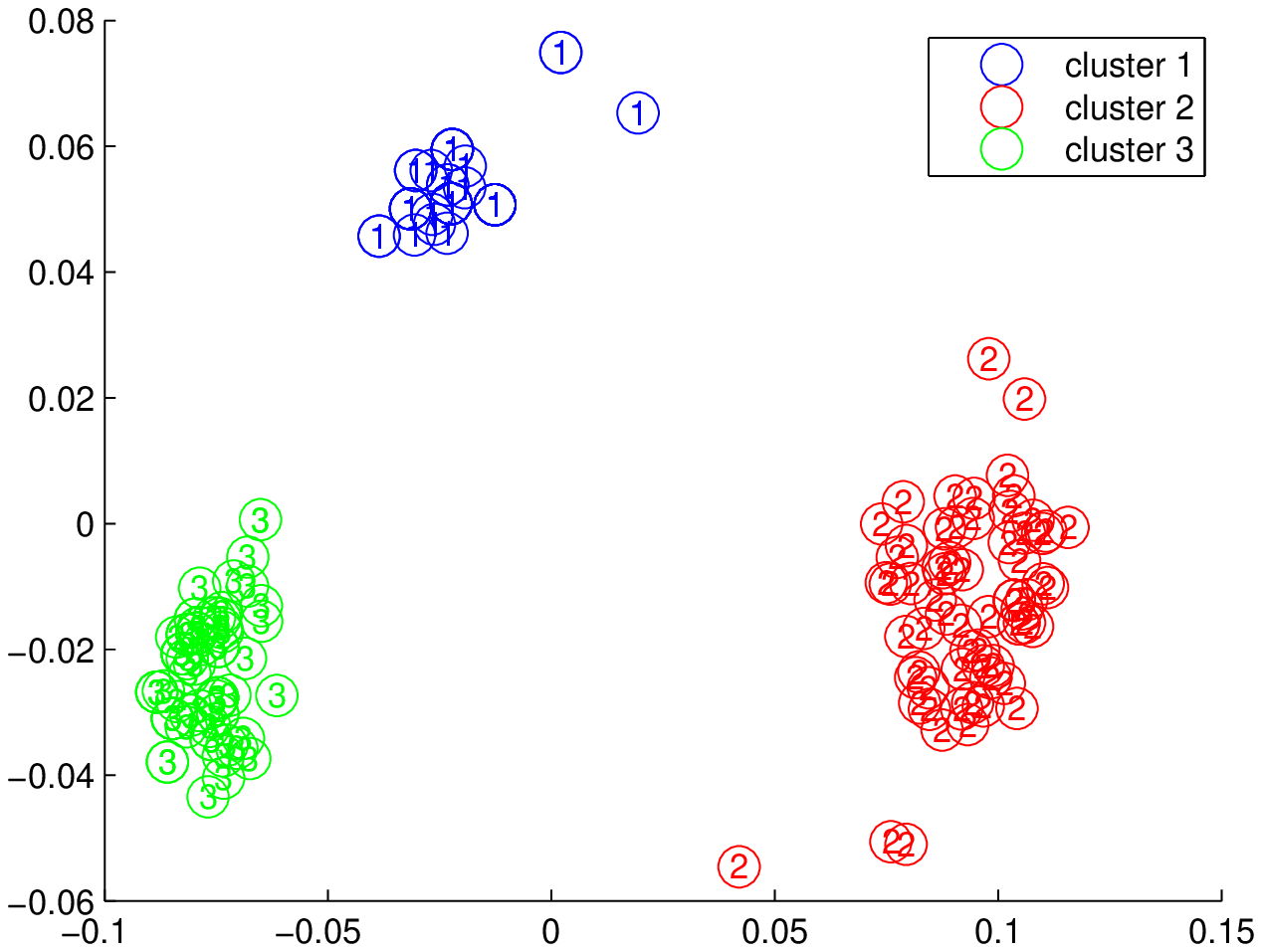}
\caption{Vectors with ground truth label projected in 2D space. The feature expectation vector is on the left and the reward vectors recovered by IRL are on the right.}
\label{fig:clusteronerule}
\end{figure}

\begin{figure}[!t]
\centering
\subfigure[Reward vectors projected in 2D space]{\includegraphics[width=2.3in]{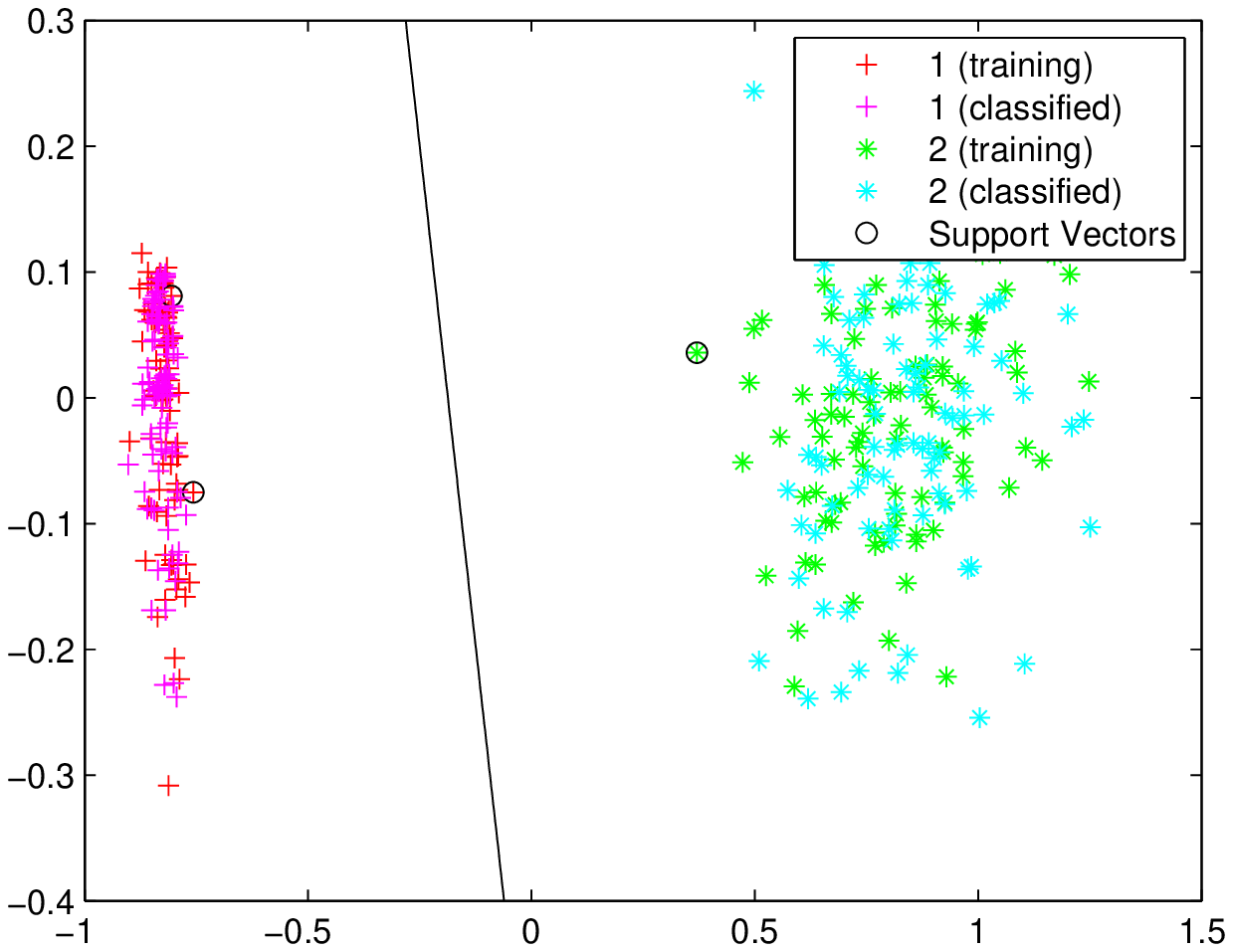}}
\hfil
\subfigure[Reward vectors projected in 3D space]{\includegraphics[width=2.3in]{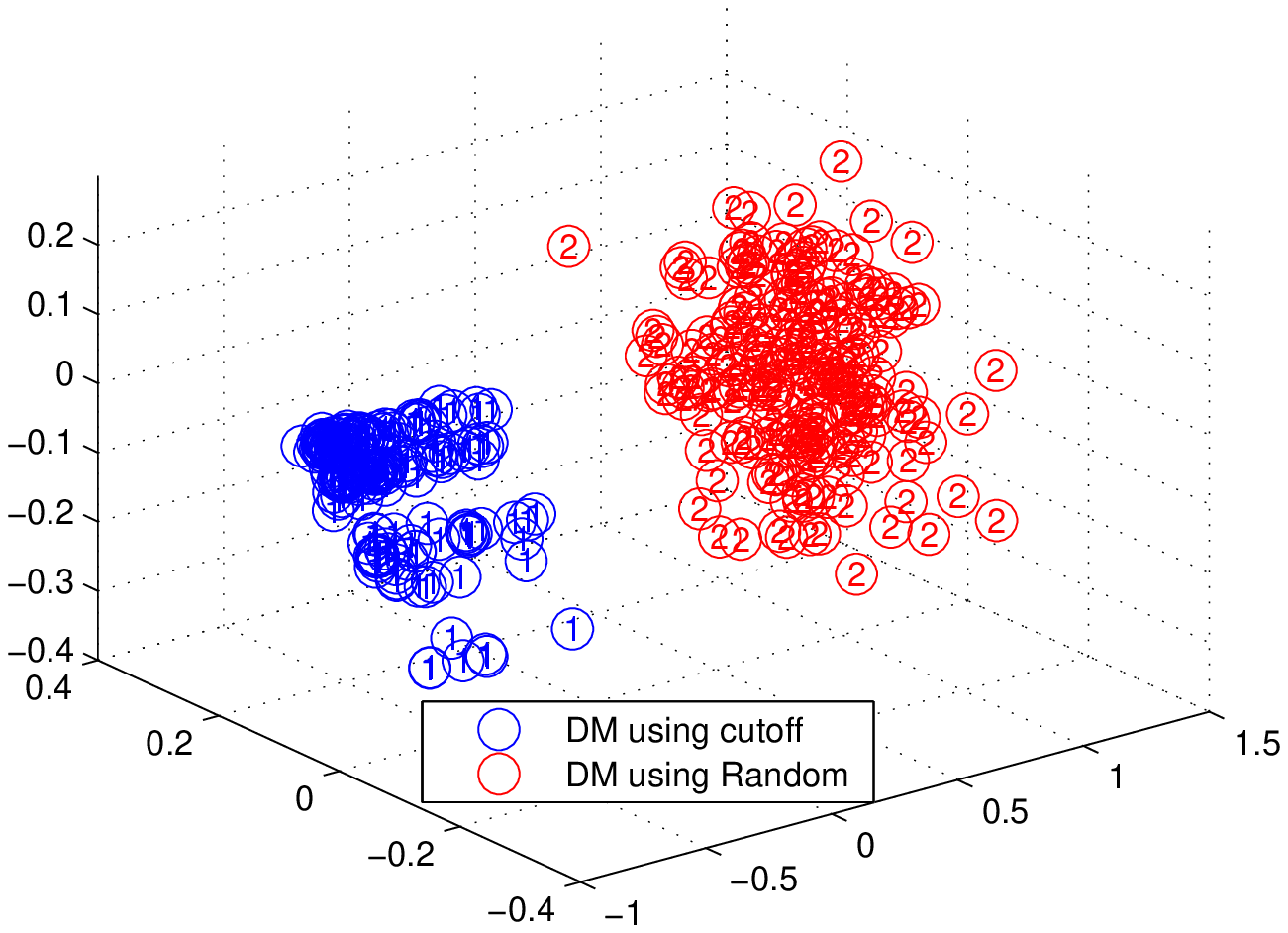}}
\caption{Visualization of a binary classification problem for subjects using cutoff rule and random rules. The \textit{PROJ} IRL algorithm is used to recover the reward vectors.}
\label{fig_classify2}
\end{figure}


\begin{table}
\begin{tabular*}{\textwidth}{@{\extracolsep{\fill}}c|cc|cc|cc}
\cline{1-7}
\multirow{2}{*} {H} &    \multicolumn{2}{c|}{CR}&    \multicolumn{2}{c|}{SNCCR}&    \multicolumn{2}{c}{CCR}\\
\cline{2-7}
 &     Action &     BIRL  &     Action &     BIRL &       Action &     BIRL     \\
\cline{1-7}
         1 &     0.0557 & {\bf 0.5497} &           0.0551 & {\bf 0.1325} &        0.0229 & {\bf 0.2081} \\

        11 &     0.3852 & {\bf 0.6893} &         0.2916 & {\bf 0.7190} &          0.1844 & {\bf 0.4974}  \\

        21 &     0.6017 & {\bf 0.7898} &         0.4305 & {\bf 0.8179} &           0.2806 & {\bf 0.5181}   \\

        31 &     0.7654 & {\bf 0.8483} &           0.5504 & {\bf 0.8641} &           0.4053 & {\bf 0.6171}   \\

        41 &     0.8356 & {\bf 0.9676} &          0.5682 & {\bf 0.9218} &         0.4524 & {\bf 0.6533}   \\

        51 &     0.8781 & {\bf 0.9739} &           0.5894 & {\bf 0.9423} &        0.5464 & {\bf 0.6507}   \\

        61 &     0.9102 & {\bf 0.9913} &         0.5984 & {\bf 0.9518} &           0.5492 & {\bf 0.6513}   \\

        71 &     0.9115 & {\bf 0.9915} &           0.6460 & {\bf 0.9639} &         0.6024 & {\bf 0.6512} \\

        81 &     0.9532 & {\bf 1.0000} &           0.6541 & {\bf 0.9721} &       {\bf 0.6708} &     0.6563   \\

        91 &     0.9707 & {\bf 1.0000} &          0.6494 & {\bf 0.9864} &       {\bf 0.6884} &     0.6544   \\
\cline{1-7}
\end{tabular*}  
\captionof{table}{NMI score for Secretary Problem}
\label{nmi_secretary_onerule}
\end{table}
Given that perfect classification performance was achieved by all algorithms, the problem of recognizing across decision rules appears to be quite easy.  A more challenging problem is to recognize variations in strategy within a single decision rule.  For each rule, we conducted recognition experiments in which 300 agents were simulated, 100 each for three distinct values  of the rule parameter.  Individuals  were differentiated by adding random noise to the parameter.   Here, we show the comparison of the clustering performance between the simple method called \textsl{FE} and our MDP model-based method.  In \figurename \ref{fig:clusteronerule}, the left figure displays an area marked ``uncertainty'' for the method called \textsl{FE}, while the right figure shows that the reward vectors have lower variance in the same group and higher variance between different groups.\figurename \ref{fig:clusteronerule} intuitively demonstrates that when the agents' behavior is represented in the reward space, the recognition problem becomes easier to solve. 

Table \ref{nmi_secretary_onerule} summarizes the  NMI scores for using K-means clustering algorithm to recognize variations in strategy within one heuristic decision rule. The column called $H$ in Table \ref{nmi_secretary_onerule} records the number of decision trajectories that have been sampled for training. Table \ref{nmi_secretary_onerule} proves that the feature representation in reward space is almost always better than the representation with statistical features computed from the raw observation data. Moreover, the reward space can particularly better characterize the behavior when the scale of the observation data is small. Note that though the MDP model cannot generate the policy that is consistent with the SNCCR and CCR rules, the reward vectors learned in the MDP environment still make the clustering problem easier to solve.

\figurename \ref{fig_classify2} shows a binary classification result of using \textit{PROJ} algorithm to learn the reward functions for the agents in Secretary problem and then categorize the agents into two groups. In this classification experiment, the users' ground truth label is either cutoff decision rule or random strategy that makes random decisions.

\section{Conclusions}
We have proposed the use of IRL to solve the problem of  behavior pattern recognition. The observed agent does not have to make decisions based on an MDP.  However, we model the agent's behavior in an MDP environment and assume that the reward function has encoded the agent's underlying decision strategies. Numerical experiments on \textsl{GridWorld} and the secretary problem suggest that the advantage that IRL enjoys over action space methods is more pronounced when observations are limited and incomplete.  We also note that there is seems to be a positive correlation between the success of IRL algorithms in apprenticeship learning (cf. \cite{qiao2011}) and their success in the behavior recognition problem.  To some degree, this relationship parallels results from \cite{konidaris12a, luis12}, where apprenticeship learning benefits from a learning structure that based on sophisticated methods for task decomposition or hierarchical identification of skill trees.

Validation of the ideas proposed here can come only through experimentation with more difficult problems.  Of particular importance would be problems involving human decision makers or other real-world scenarios, such as periodic investment, gambling, or stock trading.

\bibliography{irlnp}
\bibliographystyle{plain}
\end{document}